\pdfoutput=1
\documentclass{article}

\usepackage[nonatbib, preprint]{neurips_2023}

\usepackage{multirow}
\usepackage[utf8]{inputenc} 
\usepackage[T1]{fontenc}    
\usepackage{hyperref}       
\usepackage{url}            
\usepackage{booktabs}       
\usepackage{amsfonts}       
\usepackage{nicefrac}       
\usepackage{microtype}      
\usepackage{xcolor}         
\usepackage{times}
\usepackage{epsfig}
\usepackage{graphicx}
\usepackage{amsmath}
\usepackage{amssymb}
\usepackage{indentfirst} 
\usepackage{changepage} 
\usepackage{hyperref}
\usepackage{soul}
\usepackage{amsthm}
\usepackage{booktabs}
\usepackage{algorithm}
\usepackage{algorithmic}
\usepackage{makecell}
\usepackage{appendix}
\usepackage{subfigure}
\usepackage{fancyhdr}
\usepackage{authblk}
\usepackage{wrapfig}
\usepackage{siunitx}
\usepackage{booktabs}
\usepackage{makecell}

\usepackage{pifont}
\usepackage{fontawesome} 
\usepackage{bbding}
\usepackage{blindtext}
\newcommand*\samethanks[1][\value{footnote}]{\footnotemark[#1]}

\title{Enhancing the Performance of Transformer-based Spiking Neural Networks by SNN-optimized Downsampling with Precise Gradient Backpropagation}

\author{Chenlin Zhou$^{1}$\thanks{Equal}, Han Zhang$^{1,2}$\samethanks, Zhaokun Zhou$^{1,3}$, Liutao Yu$^{1}$, Zhengyu Ma$^{1}$\thanks{Corresponding author},  Huihui Zhou$^{1}$\samethanks, Xiaopeng Fan$^{1,2}$, Yonghong Tian$^{1,3}$\\
$^{1}$Peng Cheng Laboratory, Shenzhen 518055, China\\
$^{2}$Department of Computer Science and Technology, Harbin Institute of Technology\\
$^{3}$Department of Computer Science and Technology, Peking University\\
}

\begin{document}

\maketitle

\begin{abstract}
Deep spiking neural networks (SNNs) have drawn much attention in recent years because of their low power consumption, biological rationality and event-driven property. However, state-of-the-art deep SNNs (including Spikformer and Spikingformer) suffer from  a critical challenge related to the imprecise gradient backpropagation. This problem arises from the improper design of downsampling modules in these networks, and greatly hampering the overall model performance.
In this paper, 
we propose ConvBN-MaxPooling-LIF (CML), an SNN-optimized downsampling with precise gradient backpropagation. We prove that CML can effectively overcome the imprecision of gradient backpropagation from a theoretical perspective. In addition, we
evaluate CML on ImageNet, CIFAR10, CIFAR100, CIFAR10-DVS, DVS128-Gesture datasets. CML shows state-of-the-art performance on all these datasets in directly trained SNN models and significantly enhanced performances of transformer-based SNN models.  For instance, our model achieves 77.64 $\%$ on ImageNet, 96.04 $\%$ on CIFAR10, 81.4$\%$ on CIFAR10-DVS, with + 1.79$\%$ on ImageNet, +1.16$\%$ on CIFAR100 compared with Spikingformer. 
Codes will be available at \href{https://github.com/zhouchenlin2096/Spikingformer-CML}{Spikingformer-CML}.

\end{abstract}

\section{Introduction}
Artificial neural networks (ANNs) have demonstrated remarkable success in various artificial intelligence fields, including image classification \cite{Krizhevsky2012ImageNetCW}\cite{Simonyan2014VeryDC}\cite{Szegedy2014GoingDW}, object detection\cite{Liu2015SSDSS}\cite{Redmon2018YOLOv3AI}, and semantic segmentation\cite{Ronneberger2015UNetCN}.
Unlike ANNs, which rely on continuous high-precision floating-point data to process and transmit information, spiking neural networks (SNNs) use discrete temporal spike sequences. SNNs, as the third-generation neural network inspired by brain science\cite{maass1997networks}, have attracted the attention of many researchers in recent years due to their low power consumption, biological rationality, and event-driven characteristics. 


SNNs can be classified into two types: convolution-based SNNs and transformer-based SNNs, borrowing architectures from convolution neural network and vision transformer in ANNs, respectively.
Convolution architectures exhibit translation invariance and local dependence, but their receptive fields are typically small and limit their ability to capture global dependencies. In contrast, Vision transformer is based on self-attention mechanisms that can capture long-distance dependencies and  has enhanced the performance of artificial intelligence on many computer vision tasks, including image classification \cite{dosovitskiy2020image,yuan2021tokens}, object detection \cite{zhu2020deformable,liu2021swin} and semantic segmentation \cite{wang2021pyramid,yuan2021volo}.
Transformer-based SNNs represent a novel form of SNN that combines transformer architecture with SNN, providing great potential to break through the performance bottleneck of SNNs. So far, transformer-based SNNs mainly contain Spikformer\cite{zhou2023spikformer} and Spikingformer\cite{zhou2023spikingformer}. Spikformer introduces spike-based self-attention mechanism for the first time through spike self attention (SSA) block and shows powerful performance. However, its energy efficiency is not optimal due to its integer-float multiplications. Spikingformer, a pure event-driven transformer-based SNN, could effectively avoid non-spike computations in Spikformer through spike-driven residual learning \cite{zhou2023spikingformer}. Spikingformer significantly reduces energy consumption compared with Spikformer while even improving network performance. However, both Spikformer and Spikingformer suffer from imprecise gradient backpropagation since they inherit the traditional downsampling modules without adaption for the backpropagation of spikes. The backpropagation imprecision problem greatly limits the performance of spiking transformers.

In this paper,
we propose a downsampling module adapted to SNNs, named ConvBN-MaxPooling-LIF (CML), and prove CML can effectively overcome the imprecision of gradient backpropagation from a theoretical perspective. In addition, we evaluate CML on both static image datasets ImageNet, CIFAR10, CIFAR100 and neuromorphic datasets CIFAR10-DVS, DVS128-Gesture. 
The experimental results show our proposed CML can improve the performance of transformer-based SNNs by a large margin (e.g. + 1.79$\%$ on ImageNet, +1.16$\%$ on CIFAR100 compared with Spikingformer), and Spikingformer/Spikformer + CML achieve the state-of-the-art on all above datasets (e.g. 77.64 $\%$ on ImageNet, 96.04 $\%$ on CIFAR10, 81.4$\%$ on CIFAR10-DVS) in directly trained SNN models. 

\section{Related Work}

\textbf{Convolution-based Spiking Neural Networks.} One fundamental distinction between SNNs and traditional ANNs lies in the transmission mode of information within the networks. ANNs use continuous floating point numbers with high precision, while SNNs with spike neurons transmit information in the form of discrete temporal spike sequences. A spike neuron, the basic unit in SNNs, receives floats or spikes as inputs and accumulates membrane potential across time until it reaches a threshold to generate spikes. Typical spike neurons include Leaky Integrate-and-Fire (LIF) neuron\cite{wu2018spatio}, PLIF\cite{fang2021incorporating} and KLIF\cite{Jiang2023KLIFAO} etc. At present, there are two ways to obtain SNNs. One involves converting a pre-trained ANN to SNN\cite{bu2021optimal}\cite{Li2021AFL}\cite{Hao2023BridgingTG}, replacing the ReLU activation function in ANN with spike neurons, resulting in comparable performance to ANNs but high latency.  Another method is to directly train SNNs\cite{wu2018spatio}, using surrogate gradient\cite{neftci2019surrogate}\cite{xiao2021training} to solve the problem of non-differentiable spikes, which results in low latency but relatively poor performance. Zheng et al. propose a threshold-dependent batch normalization (tdBN) method for direct training deep SNNs\cite{zheng2021going}, which is the first time to explore the directly-trained deep SNNs with high performance on ImageNet.
Fang et al. proposed the spike-element-wise block, which overcomes problems such as gradient explosion and gradient vanishing and extended the directly trained SNNs to more than 100 layers \cite{fang2021deep}. 


\textbf{Transformer-based Spiking Neural Networks.}
Vision Transformer (ViT)\cite{dosovitskiy2020image} has become a mainstream visual network model with its superior performance in computer vision tasks. Spikeformer\cite{Li2022SpikeformerAN} has been proposed to combine the architecture of transformer to SNNs, but this algorithm still belongs to vanilla self-attention due to the existence of numerous floating-point  multiplication, division, and exponential operations. Spikformer model \cite{zhou2023spikformer}, incorporating the innovative Spiking self attention (SSA) module that effectively eliminates the complex Softmax operation in self-attention, achieves low computation, energy consumption and high performance. However, both Spikeformer and Spikformer suffer from non-spike computations (integer-float multiplications) in the convolution layer caused by the design of their residual connection. Spikingformer \cite{zhou2023spikingformer} proposes spike-driven residual learning, which could effectively avoid non-spike computations in Spikformer and Spikeformer and significantly reduce energy consumption. Spikingformer is the first pure event-driven transformer-based SNN.

\textbf{Downsamplings in SNNs.} Downsampling is a common technique used in both both SNNs and ANNs to reduce the size of feature maps, which in turn reduces the number of network parameters, accelerates the calculation speed and prevents network overfitting. Mainstream downsampling units include Maxpooling \cite{zhou2023spikformer, zhou2023spikingformer}, Avgpooling (Average pooling), and convolution with stride greater than 1\cite{fang2021deep}. Among them, Maxpooling and convolution downsampling are more commonly applied in directly trained SNNs.
In this paper, we mainly choose Maxpooling as the key downsampling unit of the proposed CML.

\section{Method}

We examine the limitation of the current downsampling technique in SNNs, which hampers the performance improvement of SNNs. We then overcome this limitation by adapting the downsampling to make it compatible with SNNs.

\subsection{Defect of Downsampling in Spikformer and Spkingformer}\label{subsec:drawbacks_of_downsampling_in_SNNs}
\begin{figure}[!t]
    \centering
    \subfigure[Downsampling in Spikformer, Spikingformer]{\includegraphics[width=0.45\linewidth]{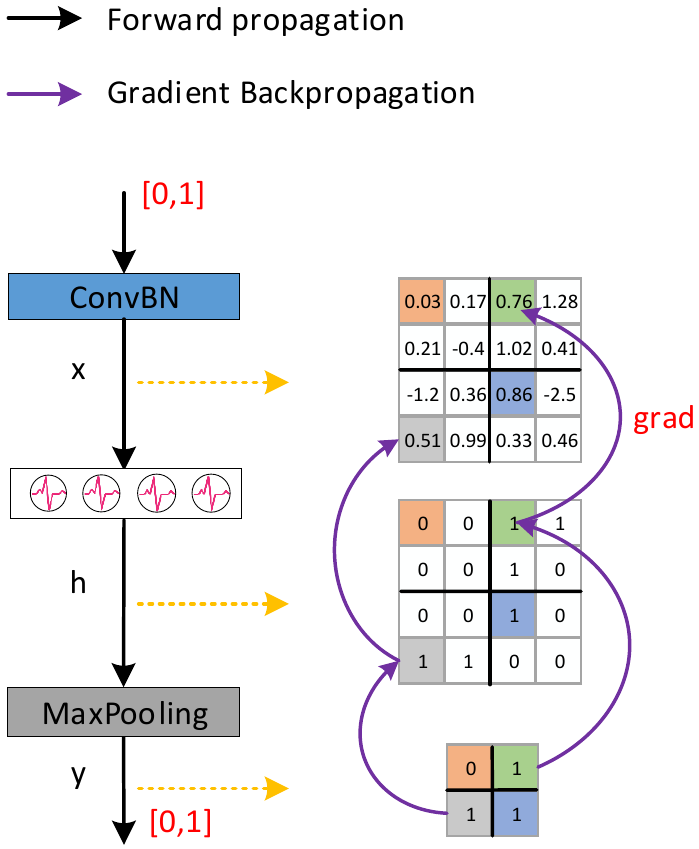}
    \label{fig:method_a}}
    \quad\quad
    \subfigure[SNN-optimized downsampling (Ours)]{\includegraphics[width=0.45\linewidth]{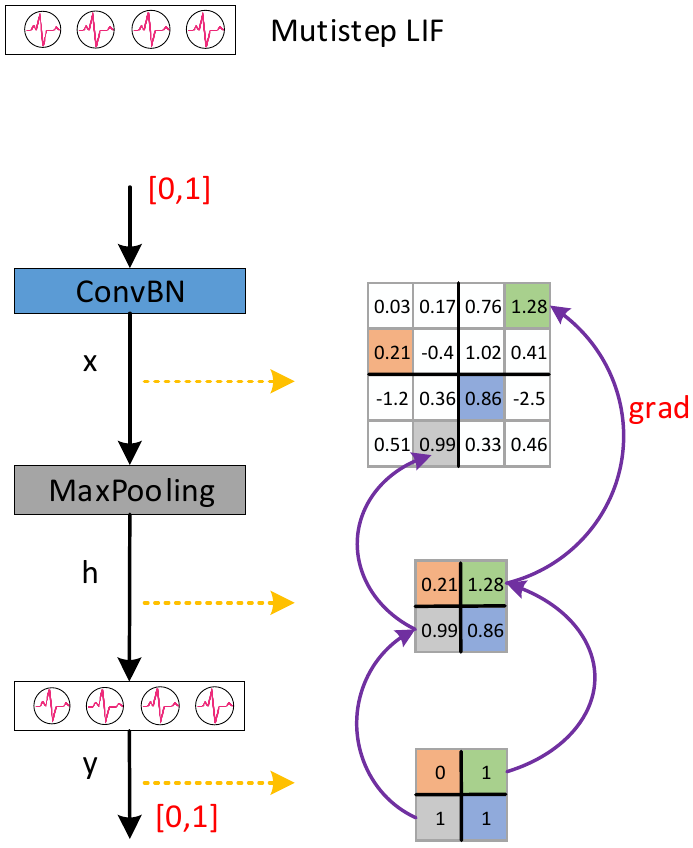}
    \label{fig:method_b}}
    \caption{The downsampling module in Spikformer, Spikingformer and our proposed SNN-optimized downsampling.
    \subref{fig:method_a} shows the downsampling module in Spikformer and Spikingformer which have imprecise gradient backpropagation issue.
    \subref{fig:method_b} shows our proposed SNN-optimized downsampling (ConvBN-MaxPooling-LIF, CML) with precise gradient backpropagation. Note that Multistep LIF is LIF spike neuron with time steps $T>1$. Same as Spikformer,  $T$ is an independent dimension for spike neuron layer. In other layers, it is merged with the batch size.
   }
    \label{fig:Neuromorphic residual}
    \vspace{-4mm}
\end{figure}

SNNs typically employ the network module shown in Figure \ref{fig:Neuromorphic residual}(a), which gives rise to the problem of imprecise gradient backpropagation. ConvBN represents the combined operation of convolution and batch normalization. Following ConvBN are the spike neurons, which receive the resultant current and accumulate the membrane potential across time, generate a spike when the membrane potential exceeds the threshold, and finally perform maxpooling for downsampling. The output of ConvBN is the feature map $x \in R^{m \times n}$, the output of spike neuron is the feature map $h \in R^{m \times n}$, and the output of maximum pooling is the feature map $y \in R^{\frac{m}{s} \times \frac{n}{s}}$, where $m \times n$ is the feature map size, and $s$ is the pooling stride.


Given the global loss function $L$ and the backpropagation gradient $\frac{\partial L}{\partial y_{i j}}$ after downsampling, the gradient at the feature map $x$ is as follows: 
\begin{equation}
    \frac{\partial L}{\partial x_{u v}}=\sum_{i=0}^{\frac{m}{s}} \sum_{j=0}^{\frac{n}{s}} \frac{\partial L}{\partial y_{i j}} \frac{\partial y_{i j}}{\partial h_{u v}} \frac{\partial h_{u v}}{\partial x_{u v}} \label{L/x_uv}
\end{equation}

The backpropagation gradient of Maxpooling is:
\begin{equation}
    \frac{\partial y_{i j}}{\partial h_{u v}}=\left\{\begin{array}{rcl}
    1,& &h_{u v}=\max \left(h_{i \times s+t, j \times s+r}\right) \\
    0,& &others
\end{array}\right.
\end{equation}

where $t,r \in [0,s)$. Suppose that the spike neuron used in our work is LIF, and the dynamic model of LIF is described as:
\begin{align}
    H[t]&=V[t-1]+\frac{1}{\tau}\left(X[t]-\left(V[t-1]-V_{\text {reset }}\right)\right)\label{H[t]}\\
    S[t]&=\Theta(H[t]-V_{th})\label{S[t]}\\
    V[t]&=H[t](1-S[t])+V_{\text {reset}}S[t]\label{V[t]}
\end{align}
where $X[t]$ is the input current at time step $t$, and $\tau$ is the membrane time constant. $V_{reset}$ represents the reset potential, $V_{th}$ represents the spike firing threshold, $H[t]$ and $V[t]$ represent the membrane potential before and after spike firing at time step $t$, respectively. $\Theta(v)$ is the Heaviside step function, if $v \geq 0$ then $\Theta(v)=1$, otherwise $\Theta(v)=0$. $S[t]$ represents the output spike at time step $t$.

The backpropagation gradient of LIF neuron is:
\begin{equation}
    \frac{\partial h_{u v}}{\partial x_{u v}}=\frac{\partial S[t]}{\partial X[t]}=\frac{1}{\tau} \times \Theta^{\prime}\left(H[t]-V[t]\right)
\end{equation}

As a result, the backpropagation gradient on feature map $x$ is:
\begin{equation}\label{eq_L/x_h}
    \dfrac{\partial L}{\partial x_{uv}}=\begin{cases}\dfrac{1}{\tau}\dfrac{\partial L}{\partial y_{ij}}*\Theta^{'}(H[t]-V[t]),& h_{uv}=\max(h_{i\times s+t,j\times s+r})\\ 0,& others\end{cases}
\end{equation}
According to Eq. (\ref{eq_L/x_h}), the gradient exists in the position of the maximal element in feature map h. However, since the output of LIF neuron are spikes, that is, the corresponding position value with spike is 1, otherwise it is 0. Therefore, the element with the first value of 1 in feature map h is chosen as the maximum value, and there is a gradient in this position, which causes imprecise gradient backpropagation. To sum up, after downsampling, when conducting backpropagation in the network structure shown in Figure \ref{fig:Neuromorphic residual}(a), the element with gradient in the feature map $x$ is not necessarily the element with the most feature information, which is the imprecision problem of gradient backpropagation.

\subsection{SNN-optimized Downsampling: CML}\label{subsec:improved_downsampling}
Here we improve the network structure of downsampling, as shown in Figure \ref{fig:Neuromorphic residual}(b), which overcomes the imprecision problem of gradient backpropagation. The output of ConvBN is the feature map $x \in R^{m \times n}$, the output of spike neuron is the feature map $h \in R^{\frac{m}{s} \times \frac{n}{s}}$, and the output of maximum pooling is the feature map $y \in R^{\frac{m}{s} \times \frac{n}{s}}$, where $m \times n$ is the feature map size, and $s$ is the pooling stride. The backpropagation gradient $\frac{\partial L}{\partial y_{i j}}$ after LIF neuron is known, then the gradient at the feature map $x$ is as follows:
\begin{equation}
    \dfrac{\partial L}{\partial x_{uv}}=\sum_{i=0}^{\frac{m}{s}}\sum_{j=0}^{\frac{n}{s}}\dfrac{\partial L}{\partial y_{ij}}\dfrac{\partial y_{ij}}{\partial h_{ij}}\dfrac{\partial h_{ij}}{\partial x_{uv}}
\end{equation}

The backpropagation gradient of Maxpooling is:
\begin{equation}
    \dfrac{\partial h_{ij}}{\partial x_{uv}}=\begin{cases}1,&x_{uv}=\max(x_{i \times s+t,j \times s+r})\\ 0,&others\end{cases}
\end{equation}
where $t,r \in [0,s)$. The backpropagation gradient of LIF neuron is:
\begin{equation}
    \frac{\partial y_{i j}}{\partial h_{i j}}=\frac{\partial S[t]}{\partial X[t]}=\frac{1}{\tau} \times \Theta^{\prime}\left(H[t]-V[t]\right)
\end{equation}
As a result, the backpropagation gradient on feature map $x$ is as follows:
\begin{equation}\label{eq_L/x_x}
    \dfrac{\partial L}{\partial x_{uv}}=\begin{cases}\dfrac{1}{\tau}\dfrac{\partial L}{\partial y_{ij}} \times \Theta^{'}(H[t]-V[t])&,x_{uv}=\max(x_{i \times s+t,j \times s+r})\\ 0&,others\end{cases}
\end{equation}
According to Eq. (\ref{eq_L/x_x}), the maximum element in feature map h corresponds to the maximum element in feature map x, that is, after downsampling, when conducting backpropagation in the network structure shown in Figure \ref{fig:Neuromorphic residual}(b), the element with gradient in feature map x is the element with the most feature information, thus overcoming the imprecision problem of gradient backpropagation. In addition, the computational cost of CML on LIF neurons is only one quarter of that downsampling in Figure \ref{fig:Neuromorphic residual}(a).

\subsection{Application and Comparative Analysis of SNN-optimized Downsampling CML }\label{subsec:application}
Our proposed SNN-optimized downsampling structure (CML) has universality in spiking neural networks. From theoretical perspectives, the downsampling structure in Figure \ref{fig:method_a} can be easily replaced by our CML structure in Figure \ref{fig:method_b} to overcome the imprecision problem of gradient backpropagation in all kinds of spiking neural networks, which improves the network performance while reduces the computational cost at the same time.
\begin{table}[tbp]
  \centering
  \caption{Experimental results of our proposed ConvBN-MaxPool-LIF Downsampling in CIFAR10/100, comparing with potential or mainstream downsampling way in SNN. In detail, we keep the retaining network structure of Spikformer and Spikingformer unchanged. Note that ConvBN-LIF-MaxPool, which is used in Spikformer and Spikingformer, is the baseline for comparation.}
    \begin{tabular}{llccc}
    \toprule
    Method & Backbone & Time Step & CIFAR10 & CIFAR100 \\
    \midrule
    ConvBN-LIF-MaxPool & Spikingformer-4-384-400E & 4     & 95.81 & 79.21 \\
    \textbf{ConvBN-MaxPool-LIF} & Spikingformer-4-384-400E & 4     & \textbf{95.95} & \textbf{80.37} \\
    ConvBN-AvgPool-LIF & Spikingformer-4-384-400E & 4     &    95.23   & 78.52 \\
    ConvBN(stride=2)-LIF & Spikingformer-4-384-400E & 4     &   94.94    & 78.65 \\
    \midrule
    ConvBN-LIF-MaxPool & Spikformer-4-384-400E & 4     & 95.51 & 78.21 \\
    \textbf{ConvBN-MaxPool-LIF} & Spikformer-4-384-400E & 4     & \textbf{96.04} & \textbf{80.02} \\
    ConvBN-AvgPool-LIF & Spikformer-4-384-400E & 4     &   95.13    & 78.53 \\
    ConvBN(stride=2)-LIF & Spikformer-4-384-400E & 4     &  94.93     & 78.02 \\
    \bottomrule
    \end{tabular}%
  \label{tab:downsamping}%
\end{table}%

In addition to the CML downsampling we proposed and ConvBN-LIF-MaxPool used in spikformer \cite{zhou2023spikformer}, we summarized another two  potential downsampling way: ConvBN-AvgPool-LIF, ConvBN(stride=2)-LIF \cite{fang2021deep} through investigation and analysis. 
Therefore, we compared these four downsampling modules by the experiment on CIFAR 10/100, which is shown in Tab. \ref{tab:downsamping}. The experimental results show our proposed ConvBN-MaxPool-LIF achieves the best performance among them, outperforming others by a large margin. In Sec.\ref{sec:exp}, we carry out extensive experiments to further verify the effectiveness of our SNN-optimized downsampling module.

\section{Experiments} \label{sec:exp} 
In this section, we evaluate the CML downsampling module on static datasets(ImageNet\cite{deng2009imagenet}, CIFAR10, and CIFAR100\cite{krizhevsky2009learning}) and neuromorphic datasets (CIFAR10-DVS and DVS128 Gesture \cite{amir2017dvsg}), using Spikformer and Spikingformer as baselines. Spikformer and Spikingformer are representative transformer-based spiking neural networks.
Specifically, we replace SPS with CML in spikformer and replace SPED with CML in spikingformer, while  keeping the remaining settings unchanged.

\subsection{ImageNet Classification}\label{sec:imagenet} 
\textbf{ImageNet-1K} contains around $1.3$ million $1000$-class images for training and $50,000$ images for validation. We conduct experiments on ImageNet-1K to evaluate our CML module, with an input size of $224\times 224$ by default both during training and inference. The training details of our proposed Spikformer + CML and Spikingformer + CML remain consistent with the original Spikformer and Spikingformer, respectively. 

The experimental results shown in Tab. \ref{tab:imagenet} indicate that CML significantly enhances the performance of Spikformer and Spikingformer with various network sizes. Specifically, Spikformer-8-768 + CML achieves 76.55$\%$ Top-1 classification accuracy, which outperforms Spikformer-8-768 by 1.74$\%$. In addition, Spikingformer-8-768 + CML achieves 77.64$\%$ Top-1 classification accuracy, which outperforms Spikingformer-8-768 by 1.79$\%$  and achieves the state-of-the-art performance on ImageNet in directly trained spiking neural network models. These results strongly validate the effectiveness of CML.
\begin{table}[!tbp]
  \centering
  \caption{Evaluation on ImageNet. The default input resolution of all the models in inference is 224 $\times$ 224. CML module enhances the network performance of all models of Spikformer and Spikingformer by a large margin.
  Note that 77.64 $\%$ of Spikingformer-8-768 + CML achieves the state-of-the-art performance on ImageNet in directly trained SNN models. }
    \begin{tabular}{p{3.8cm}<{\raggedright}p{3.8cm}<{\raggedright}p{1.2cm}<{\centering}p{1.0cm}<{\centering}p{2.0cm}<{\centering}}
    \toprule
    \multicolumn{1}{l}{Methods} & \multicolumn{1}{l}{Architecture} & Param\newline{}(M)  & {Time Step} & {Top-1 Acc} \\
    \midrule
    Hybrid training\cite{rathi2020enabling} & ResNet-34 &21.79  &250 &61.48\\
    {\multirow{2}{*}{TET\cite{deng2021temporal}}} & Spiking-ResNet-34 &21.79  &6 & {64.79} \\
                                                                     & SEW-ResNet-34 &21.79  &4 & {68.00} \\
    {\multirow{2}{*}{Spiking ResNet\cite{hu2018residual}}} &ResNet-34 &21.79  & 350 & 71.61 \\
                                                                              &ResNet-50 &25.56 & 350 & 72.75 \\
    \multicolumn{1}{l}{\multirow{1}{*}{STBP-tdBN\cite{zheng2021going}}} &Spiking-ResNet-34 &21.79  & 6 & 63.72 \\
    \multirow{4}{*}{SEW ResNet\cite{fang2021deep}}  & SEW-ResNet-34 &21.79  &4 & 67.04 \\
                                                     & SEW-ResNet-50 &25.56   &4 & 67.78 \\
                                                     & SEW-ResNet-101 &44.55   &4 & 68.76 \\
                                                     & SEW-ResNet-152 &60.19   &4 & 69.26 \\
    MS-ResNet\cite{hu2021advancing} & ResNet-104 &44.55+ &5 &74.21\\
    {Transformer (ANN)\cite{zhou2023spikformer}}   & {Transformer-8-512} & {29.68}  & {1} & {{80.80}} \\
    \midrule
    {\multirow{3}{*}{{Spikformer\cite{zhou2023spikformer}}}}&{Spikformer-8-384} &16.81 &4 & 70.24\\
                                   &{Spikformer-8-512}&29.68 & 4 & {73.38}\\
                                   &{Spikformer-8-768}&66.34 & 4 & {74.81}\\

    \midrule
    {\multirow{3}{*}{{\textbf{Spikformer + CML}}}}&{Spikformer-8-384} &16.81 &4 & \textbf{72.73(+2.49)}\\
                                   &{Spikformer-8-512}&29.68 & 4 & \textbf{75.61(+2.23)}\\
                                   &{Spikformer-8-768}&66.34 & 4 & \textbf{77.34(+2.53)}\\

    \midrule
    {\multirow{3}{*}{Spikingformer\cite{zhou2023spikingformer}}}&{Spikingformer-8-384} &16.81 &4 & 72.45\\
                                   &{Spikingformer-8-512}&29.68 & 4 & 74.79\\
                                   &{Spikingformer-8-768}&66.34 & 4 & {75.85}\\
    \midrule
    {\multirow{3}{*}{\textbf{Spikingformer + CML}}}&{Spikingformer-8-384} &16.81 &4 & \textbf{74.35(+1.90)}\\
                                   &{Spikingformer-8-512}&29.68 & 4 & \textbf{76.54(+1.75)}\\
                                   &{Spikingformer-8-768}&66.34 & 4 & {\textbf{77.64(+1.79)}}\\
    
    \bottomrule
    \end{tabular}%
  \label{tab:imagenet}%
\end{table}%

\subsection{CIFAR Classification}\label{sec:cifar}
\textbf{CIFAR10/CIFAR100} both contain 50,000 train and 10,000 test images with 32 × 32 resolution. CIFAR10 and CIFAR100 contain 10 categories and 100 categories for classification, respectively. We evaluate our CML module on CIFAR10 and CIFAR100. The training details of our proposed Spikformer + CML and Spikingformer + CML are consistent with the original Spikformer and Spikingformer, respectively.

The experimental results are shown in Tab. \ref{tab:cifar}. CML enhances the performance of all Spikformer and Spikingformer models in both CIFAR10 and CIFAR100 by a large margin. For CIFAR10,
Spikformer-4-384-400E + CML achieves 96.04$\%$ Top-1 classification accuracy, which outperforms Spikformer-4-384-400E by 0.53$\%$  and realizes the state-of-the-art performance of CIFAR10 in directly trained  spiking neural network model.
Spikingformer-4-384-400E + CML achieves 95.95$\%$ Top-1 classification accuracy, which outperforms Spikingformer-4-384-400E by 0.14$\%$.
For CIFAR100,
Spikformer-4-384-400E + CML achieves 80.02$\%$ Top-1 classification accuracy, which outperforms Spikingformer-4-384-400E by 1.81$\%$.
Spikingformer-4-384-400E + CML achieves 80.37$\%$ Top-1 classification accuracy, which outperforms Spikingformer-4-384-400E by 1.16$\%$  and realizes the state-of-the-art performance of CIFAR100 in directly trained  spiking neural network model.  The experimental results strongly further verify the effectiveness of our method.

\begin{table}[!tbp]
  \centering
  \caption{Performance comparison of our method on CIFAR10/100. CML module enhances the network performance of all models of Spikformer and Spikingformer in both CIFAR10 and CIFAR100 by a large margin. Note that 96.04 $\%$ of Spikformer-4-384-400E + CML and 80.37 $\%$ of Spikingformer-4-384-400E + CML are the state-of-the-art performance of CIFAR10 and CIFAR100 in directly trained  spiking neural networks, respectively.
  }
    \begin{tabular}{p{3.0cm}<{\raggedright}p{4.0cm}<{\centering}p{0.6cm}<{\centering}p{0.5cm}<{\centering}p{1.6cm}<{\centering}p{1.5cm}<{\centering}}
    \toprule
    Methods & Architecture & Param\newline{}(M) & Time Step & CIFAR10 Acc & CIFAR100 Acc \\
   \midrule
    Hybrid training\cite{rathi2020enabling} &VGG-11 &9.27 &125 &92.22 &67.87\\
    Diet-SNN\cite{rathi2020diet} &ResNet-20 &0.27 &10\textbf{/}5  & 92.54& 64.07\\
    STBP\cite{wu2018spatio} &CIFARNet &17.54&12 & 89.83&-\\
    STBP NeuNorm\cite{wu2019direct} &CIFARNet &17.54 &12 &90.53& -\\
    TSSL-BP\cite{zhang2020temporal} &CIFARNet &17.54 &5 &91.41& -\\
    STBP-tdBN\cite{zheng2021going} &ResNet-19 &12.63 & 4 & 92.92 & 70.86\\
    TET\cite{deng2021temporal}  &\multicolumn{1}{c}{\multirow{1}{*}{ResNet-19}} &12.63 & 4 & {94.44}& {74.47}\\
    \multicolumn{1}{l}{\multirow{2}{*}{{MS-ResNet\cite{hu2021advancing}}}} &ResNet-110 &- & - & 91.72 & 66.83\\
                                     &ResNet-482 &- & - & 91.90 & -\\
   \midrule
   \multicolumn{1}{l}{\multirow{2}{*}{{ANN\cite{zhou2023spikformer}}}}  &ResNet-19* &12.63 &1 & 94.97 & 75.35\\
                                                      & {Transformer-4-384} & {9.32} &1 & 96.73 & 81.02 \\
    \midrule
    \multicolumn{1}{l}{\multirow{4}{*}{Spikformer\cite{zhou2023spikformer}}}  & Spikformer-4-256 &4.15 & 4 & 93.94 & {75.96}\\
                                                       & {Spikformer-2-384} &5.76 & 4 & {94.80} & {76.95}\\
                                                       & {Spikformer-4-384} &9.32 & 4 & {95.19} & {77.86}\\
                                                       & {Spikformer-4-384-400E} &9.32 & 4 & {95.51} & {78.21}\\

    \midrule
    \multicolumn{1}{l}{\multirow{4}{*}{\textbf{Spikformer+CML}}}  & Spikformer-4-256 &4.15 & 4 & \textbf{94.82(+0.88)} & \textbf{77.64(+1.68)}\\
                                                       & {Spikformer-2-384} &5.76 & 4 & \textbf{95.63(+0.83)} & \textbf{78.75(+1.80)}\\
                                                       & {Spikformer-4-384} &9.32 & 4 & \textbf{95.93(+0.74)} & \textbf{79.65(+1.79)}\\
                                                       & {Spikformer-4-384-400E} &9.32 & 4 & \textbf{96.04(+0.53)} & \textbf{80.02(+1.81)}\\
                                                       
    \midrule
    \multicolumn{1}{l}{\multirow{4}{*}{Spikingformer\cite{zhou2023spikingformer}}}  
                                                       & Spikingformer-4-256 &4.15 & 4 & {94.77} & {77.43}\\
                                                       & {Spikingformer-2-384} &5.76 & 4 & {95.22} & {78.34}\\
                                                       & {Spikingformer-4-384} &9.32 & 4 & {95.61} & {79.09}\\
                                                       & {Spikingformer-4-384-400E} &9.32 & 4 & {95.81} & {79.21}\\

    \midrule
    \multicolumn{1}{l}{\multirow{4}{*}{\textbf{Spikingformer+CML}}} 
                                                       & Spikingformer-4-256 &4.15 & 4 & \textbf{94.94(+0.17)} & \textbf{78.19(+0.76)}\\
                                                       & {Spikingformer-2-384} &5.76 & 4 & \textbf{95.54(+0.32)} & \textbf{78.87(+0.53)}\\
                                                       & {Spikingformer-4-384} &9.32 & 4 & \textbf{95.81(+0.20)} & \textbf{79.98(+0.89)}\\
                                                       & {Spikingformer-4-384-400E} &9.32 & 4 & \textbf{95.95(+0.14)} & \textbf{80.37(+1.16)}\\
    \bottomrule
    \end{tabular}%
  \label{tab:cifar}%
\end{table}%

\subsection{DVS Classification}\label{sec:dvs}
\textbf{CIFAR10-DVS Classification.} 
 CIFAR10-DVS is a neuromorphic dataset derived from the CIFAR10 dataset, where the visual input is captured by a Dynamic Vision Sensor (DVS) that represents changes in pixel intensity as asynchronous events rather than static frames. It includes 9,000 training samples and 1,000 test samples. We carry out experiments on CIFAR10-DVS to evaluate our CML module. The training details of our proposed Spikformer + CML and Spikingformer + CML are consistent with the original Spikformer and Spikingformer, which all contain 2 spiking transformer blocks with 256 patch embedding dimensions. 

We compare our method with SOTA methods on CIFAR10-DVS in Tab.\ref{tab:dvs}. Spikingformer + CML achieves 81.4$\%$ top-1 accuracy with 16 time steps and 80.5$\%$ accuracy with 10 time steps, outperforming Spikingformer by 0.1$\%$ and 0.6$\%$ respectively. Spikformer + CML achieves 80.9$\%$ top-1 accuracy with 16 time steps and 79.2$\%$ accuracy with 10 time steps, outperforms Spikformer by 0.3$\%$ and 0.6$\%$ respectively. Among them, 81.4$\%$ of Spikingformer + CML is the state-of-the-art performance of CIFAR10-DVS in directly trained spiking neural network.

\textbf{DVS128 Gesture Classification.} DVS128 Gesture is a gesture recognition dataset that contains 11 hand gesture categories from 29 individuals under 3 illumination conditions. The training details of our proposed Spikformer + CML and Spikingformer + CML are consistent with the original Spikformer and Spikingformer on DVS128 Gesture, which all contain 2 spiking transformer blocks with 256 patch embedding dimensions. 

We compare our method with SOTA methods on DVS128 Gesture in Tab.\ref{tab:dvs}. Spikingformer + CML achieves 98.6$\%$ top-1 accuracy with 16 time steps and 97.2$\%$ accuracy with 10 time steps, outperforms Spikingformer by 0.3$\%$ and 1.0$\%$ respectively. Spikformer + CML achieves 98.6$\%$ top-1 accuracy with 16 time steps and 97.6$\%$ accuracy with 10 time steps, outperforms Spikformer by 0.7$\%$ and 1.8$\%$ respectively. Among them, 98.6$\%$ of Spikingformer + CML and Spikformer + CML is the state-of-the-art performance of DVS128 Gesture in  directly trained  spiking neural network.

\begin{table}[!tbp]
  \centering
  \caption{Results on two neuromorphic datasets, CIFAR10-DVS and DVS128 Gesture. The result of Spikformer is our implementation according to its open source code. Note that 81.4 $\%$ and 98.6 $\%$ of Spikingformer + CML achieve the state-of-the-art performance of CIFAR10-DVS and DVS128 Gesture in directly trained  spiking neural networks, respectively.}
    \begin{tabular}{lp{1.8cm}<{\centering}p{1.8cm}<{\centering}p{1.8cm}<{\centering}p{1.8cm}<{\centering}}
    \toprule
    \multirow{2}[4]{*}{Method}  & \multicolumn{2}{c}{CIFAR10-DVS} & \multicolumn{2}{c}{DVS128} \\
    \cmidrule{2-5}               & Time Step     & \multicolumn{1}{c}{Acc} & Time Step & \multicolumn{1}{c}{Acc} \\
    \midrule
    LIAF-Net \cite{wu2021liaf}      & 10 & {70.4} & {60} & {97.6} \\
    TA-SNN \cite{yao2021temporal}    & 10 & {72.0} & {60} & {98.6} \\
    Rollout \cite{kugele2020efficient}   & 48 & {66.8} & {240} & {97.2} \\
    DECOLLE \cite{kaiser2020synaptic}    & -  & {-} & {500} & {95.5} \\
    tdBN \cite{zheng2021going}           & 10 & {67.8} & {40} & {96.9} \\
    PLIF \cite{fang2021incorporating}     & 20 & {74.8} & {20} & {97.6} \\
    SEW-ResNet \cite{fang2021deep}     & 16 & {74.4} & {16} & {97.9} \\
    Dspike \cite{li2021differentiable}   & 10 & {75.4} & {-} & {-} \\
    SALT \cite{kim2021optimizing}    & 20 & {67.1} & {-} & {-} \\
    DSR \cite{meng2022training}            & 10 & {77.3} & {-} & {-} \\
    MS-ResNet \cite{hu2021advancing}            & - & {75.6} & {-} & {-} \\
    \midrule
    \multirow{2}{*}{Spikformer\cite{zhou2023spikformer} (Our Implement)}      & 10    & {78.6} & {10} & {95.8} \\
                                       & 16    & {80.6} & {16} & {97.9} \\
    \midrule
    \multirow{2}{*}{\textbf{Spikformer + CML}}      & 10    & \textbf{79.2(+0.6)} & {10} & \textbf{97.6(+1.8)} \\
                                       & 16    & \textbf{80.9(+0.3)} & {16} & \textbf{98.6(+0.7)} \\
    \midrule
    \multirow{2}{*}{Spikingformer\cite{zhou2023spikingformer}}   & 10   & 79.9  & 10   & {96.2} \\
                                         & 16   &  {81.3}  &  16  & {98.3} \\
    \midrule
    \multirow{2}{*}{\textbf{Spikingformer + CML}}   & 10   & \textbf{80.5(+0.6)}  & 10   & \textbf{97.2(+1.0)} \\
                                         & 16   &  \textbf{81.4(+0.1)}  &  16  & \textbf{98.6(+0.3)} \\
    \bottomrule
    \end{tabular}%
  \label{tab:dvs}%
\end{table}%


\section{Conclusion}
In this paper, we investigate the imprecision problem of gradient backpropagation caused by downsampling in Spikformer and Spikingformer. Subsequently, we propose an SNN-optimized downsampling module with precise gradient backpropagation, named ConvBN-MaxPooling-LIF (CML), and prove our proposed CML can effectively overcome the imprecision of gradient backpropagation from theoretical perspectives. In addition, we
evaluate CML on the static datasets ImageNet, CIFAR10, CIFAR100 and neuromorphic datasets CIFAR10-DVS, DVS128-Gesture. The experimental results show our proposed CML can improve the performance of SNNs by a large margin (e.g. + 1.79$\%$ on ImageNet, +1.16$\%$ on CIFAR100 comparing with Spikingformer), and our models achieve the state-of-the-art on all above datasets (e.g. 77.64 $\%$ on ImageNet, 96.04 $\%$ on CIFAR10, 81.4$\%$ on CIFAR10-DVS) in directly trained SNNs.

\section{Acknowledgment}

This work is supported by grants from National Natural Science Foundation of China 62236009 and 62206141.

\bibliographystyle{unsrt}
\bibliography{root}

\section*{Appendix}

\appendix
\section{Additional Results}
\subsection{Additional Results on CIFAR10}
We trained Spikingformer + CML on CIFAR10 up to 600 epochs, and the accuracy could increase up to 96.14$\%$. 
\begin{table}[htbp]
  \centering
  \caption{Training Spikingformer up to 600 epochs on CIFAR10.}
    \begin{tabular}{cccc}
    \toprule
    Backbone & models & Timestep & CIFAR10 \\
    \midrule
    \multicolumn{1}{l}{\multirow{3}{*}{\textbf{Spikingformer+CML}}}
    &Spikingformer-4-384-300E & 4     & 95.81 \\
    &Spikingformer-4-384-400E & 4     & 95.95 \\
    &Spikingformer-4-384-600E & 4     & \textbf{96.14} \\
    \bottomrule
    \end{tabular}%
  \label{tab:cifar600e}%
\end{table}%

\subsection{Loss and Accuracy when training on CIFAR100}
\begin{figure}[htbp]
    \center
    \subfigure[Training loss]{
    \begin{minipage}[c]{0.31\linewidth}
    \centering
    \includegraphics[width=1.0\linewidth]{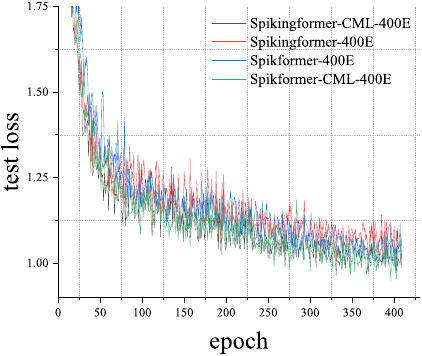}
    \end{minipage}
    }
    \subfigure[Testing loss]{
    \begin{minipage}[c]{0.31\linewidth}
    \centering
    \includegraphics[width=1.0\linewidth]{figs/c100-testloss.pdf}
    \end{minipage}
    }
    \subfigure[Test accuracy]{
    \begin{minipage}[c]{0.31\linewidth}
    \centering
    \includegraphics[width=1.0\linewidth]{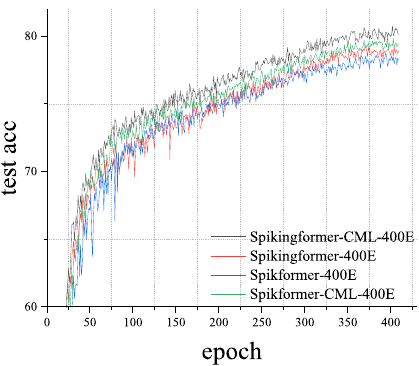}
    \end{minipage}
    }
\caption{Comparison of the training loss, testing loss and test accuracy on CIFAR100 dataset.}
\label{fig:loss&acc}
\end{figure}
Fig. \ref{fig:loss&acc} visualizes the training loss, testing loss and test accuracy of Spikingformer + CML, Spikingformer, Spikformer + CML, Spikformer on CIFAR100 dataset, respectively. All models have been trained with 400 epochs. The results further verify  the effectiveness of our SNN-optimized downsampling CML. 

\subsection{Supplement of experimental details}
In our experiments, we use 8 GPUs when training on ImageNet, while 1 GPU is used to train other four datasets. In addition, we 
adjust the value of membrane time constant $\tau$ in spike neuron when training models on DVS datasets. When directly training SNN models with surrogate function, 
\begin{equation}
{Sigmoid}(x)=\frac{1}{1+\exp (-\alpha x)}
\end{equation}
we select the Sigmoid function as the surrogate function with $\alpha=4$ in all experiments.
\end{document}